\documentclass[10pt,journal,compsoc]{IEEEtran}

\ifCLASSOPTIONcompsoc
  \usepackage[nocompress]{cite}
\else
  \usepackage{cite}
\fi

\def\BreakURLText#1{\@tfor\brk@tempa:=#1\do{\brk@tempa\hskip0pt}}
\let\lt=<
\let\gt=>
\def\processVert{\ifmmode|\else\textbar\fi}

\usepackage{hyperref}

\ifCLASSINFOpdf

\else

\fi

\usepackage{graphicx}
\graphicspath{ {./images/} }
\usepackage{subcaption}

\def\cAlignHack{\rightskip\@flushglue\leftskip\@flushglue\parindent\z@\parfillskip\z@skip}
\def\rAlignHack{\rightskip\z@skip\leftskip\@flushglue \parindent\z@\parfillskip\z@skip}

\usepackage{tabulary,xcolor}
\usepackage{amsfonts,amsmath,amssymb}
\usepackage[T1]{fontenc}

\usepackage{url,multirow,morefloats,floatflt,cancel,tfrupee}

\def\ps@pprintTitle{\save@ps@pprintTitle\gdef\@oddfoot{\footnotesize\itshape \null\hfill\today}}
\def\hlinewd#1{%
  \noalign{\ifnum0=`}\fi\hrule \@height #1%
  \futurelet\reserved@a\@xhline}

\usepackage{tabulary,xcolor}
\usepackage{amsfonts,amsmath,amssymb}
\usepackage[T1]{fontenc}
\makeatletter

\usepackage{amsmath} 

\newif\ifmultipleabstract\multipleabstractfalse%
\begin{document}
%
\title{A Novel Approach for Robust Multi Human Action Recognition and Summarization based on 3D Convolutional Neural Networks }

\author{Noor Almaadeed,{}
        Omar Elharrouss,{}
        Somaya Al-Maadeed,{}
        Ahmed Bouridane,{}%
	Azeddine Beghdadi{}%
\thanks{Noor Almaadeed, Omar Elharrouss, Somaya Al-Maadeed are affiliated with the Department of Computer Science and Engineering, Qatar University,Doha, Qatar e-mail: (n.alali@qu.edu.qa, elharrouss.omar@gmail.com, and S\_alali@qu.edu.qa).

Ahmed Bouridane is affiliated with Department of Computer and Information Sciences, Northumbria University, Newcastle upon Tyne NE1 8ST, United Kingdom.

Azeddine Beghdadi is affiliated with Department of Computer and Information Sciences, Galilee Institute,Paris 13 Nord Univeristy, Paris, France.}
}

\markboth{}%
{Shell \MakeLowercase{\textit{et al.}}: Bare Demo of IEEEtran.cls for Computer Society Journals}


\IEEEtitleabstractindextext{%
\begin{abstract}
Human actions in videos are 3D signals. However, there are a few methods available for multiple human action recognition. For long videos, it's difficult to search within a video for a specific action and/or person. For that, this paper proposes a new technic for multiple human action recognition and summarization for surveillance videos. The proposed approach proposes a new representation of the data by extracting the sequence of each person from the scene. This is followed by an analysis of each sequence to detect and recognize the corresponding actions using 3D convolutional neural networks (3DCNNs). Action-based video summarization is performed by saving each person's action at each time of the video. Results of this work revealed that the proposed method provides accurate multi human action recognition that easily used for summarization of any action. Further, for other videos that can be collected from the internet, which are complex and not built for surveillance applications, the proposed model was evaluated on some datasets like UCF101 and YouTube without any preprocessing. For this category of videos, the summarization is performed on the video sequences by summarizing the actions in each subsequence. The results obtained demonstrate its efficiency compared to state-of-the-art methods. 

\end{abstract}

\begin{IEEEkeywords}
Action recognition, multi human action recognition, video summarization, 3-dimensional convolutional neural networks.
\end{IEEEkeywords}}

\maketitle

\IEEEdisplaynontitleabstractindextext

%
\IEEEpeerreviewmaketitle

\IEEEraisesectionheading{\section{Introduction}\label{sec:introduction}}

%
\IEEEPARstart{A} large number of videos that capture different events throughout the world are produced on a daily basis using different types of cameras (surveillance, phones, and filming crew). However, detecting the temporal and spatial events requires retrieving the required information contained from the captured videos. This process of creating useful information about the content of videos can be achieved through video content analysis \cite{1} which has a wide range of applications including the retrieval of particular information, providing automatic alerts, and data summarization. Human action recognition is another specific application of video content analysis which aims to recognize activities from a series of observations on actions of subjects and the surrounded environment and it is important for many other applications \cite{2}. However, human action recognition is a complex technology due to the difficulty to extract information about a person's identity and psychological states. Hence; human action recognition has gained the interest of researchers in the computer vision area and many approaches have been proposed for the aim of developing this technology.

\begin{figure}[t!]
  \centering
  \begin{subfigure}[b]{0.46\linewidth}
    \includegraphics[width=\linewidth]{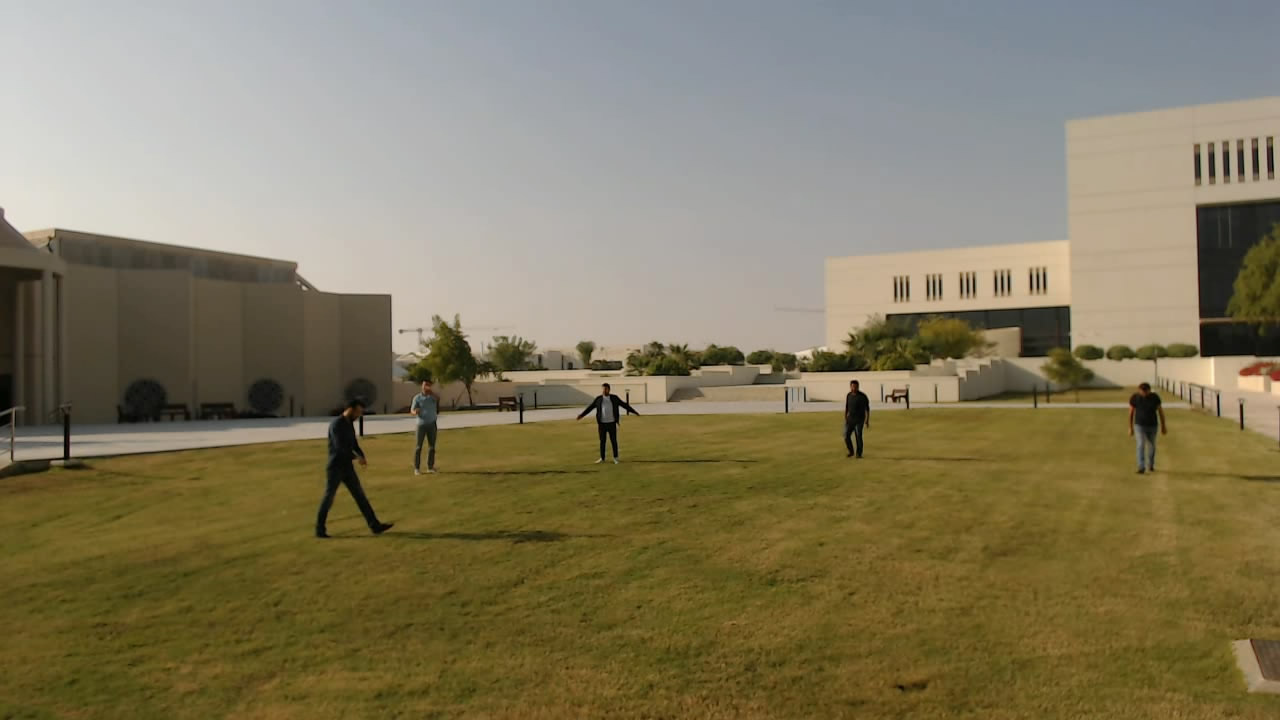}
    \caption{}
  \end{subfigure}
  \begin{subfigure}[b]{0.46\linewidth}
    \includegraphics[width=\linewidth]{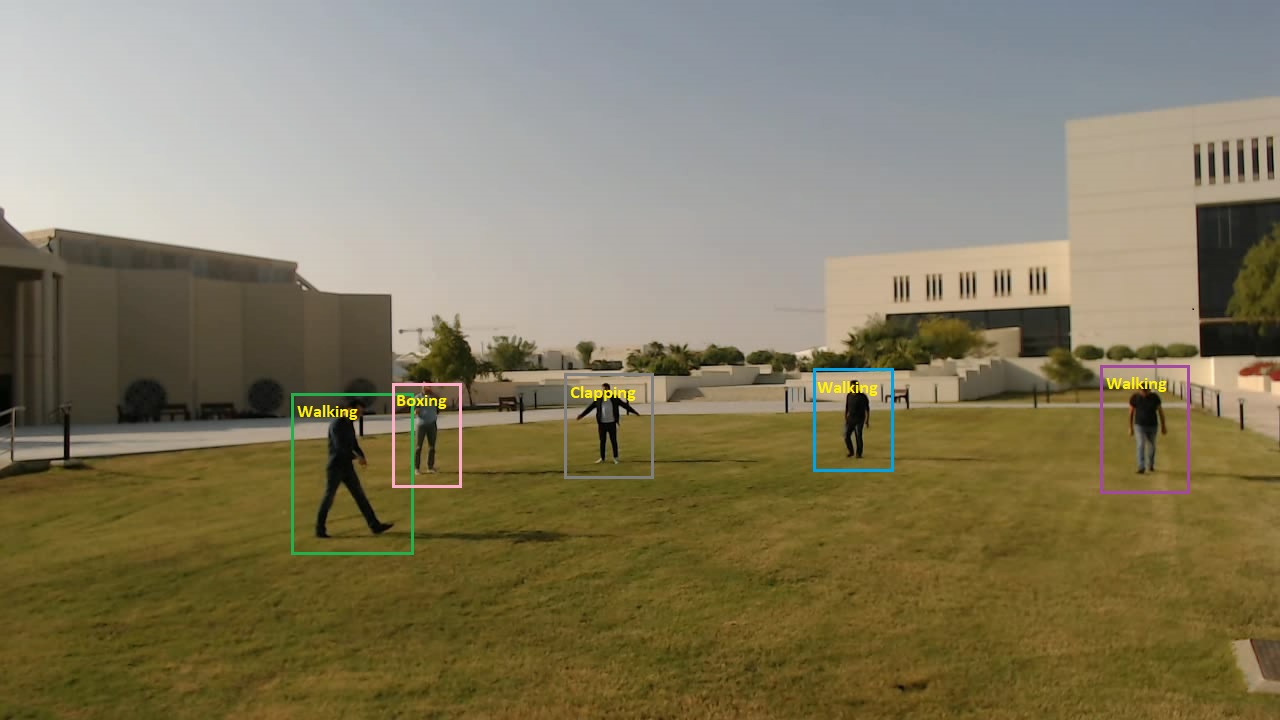}
    \caption{}
  \end{subfigure}

  \caption{Multiple huamn action recognition example. (a) frame of a video. (b) detected and recognized actions in the video.}
  \label{fig:g}
\end{figure}

Human action recognition is an important task for many applications including video surveillance systems, video indexing and retrieval, sport application, and multimedia. Action detection, recognition, and summarization can be exploited to support many other tasks. For example, in sports applications, it can help to recognize and understand players' poses allowing to make a good decision on players' fouls, especially for football games that require correct decisions in many situations. Each human action has many characteristic patterns and appearances. In video processing, the effectiveness of such an action recognition method is related to the chosen features. Therefore, in order to recognize actions from video sequences, the existing algorithms available in the literature combine image action and spatiotemporal features, such that, the spatial features represent the visual appearance while the temporal one illustrates the dynamic motion. 

{\renewcommand{\arraystretch}{1.3}%
\begin{table*}[t!]
\caption{Human action recognition methods description}
\centering
\small
\begin{tabular}{ p{2 cm}  p{6.7cm}   p{3.5cm}  p{2cm} }
\hline
\textbf{Methods} & \textbf{Description} & \textbf{Videos} & \textbf{Datasets} \\
\hline
Detection-Recognition & -Detection of the ROI of the human body before the recognition in this region. \newline -Easy extraction of motion features such as MHI, HOF. &
  -Static background  -Fixed camera & 
KTH\newline IXMAS \newline UCF-ARG \newline PETS \\  \cline{1-4} 

Video  Classification  &
  -Classification of videos according to the content in it the videos.\newline -Independent of the number of actors in the video \newline Related to the appearance of action in the video. &
  -Dynamic background \newline -Moving camera &
  Hollywood \newline Hollywood2 \newline UCF Sport  \newline UCF50-101 \newline HMDB51\\
\hline
\end{tabular}

\end{table*}\quad
Human action recognition approaches can be classified into two main categories: detection recognition and video classification. The detection recognition approaches start by first detecting the motion of the person's action, followed by recognizing the action(s). These approaches are validated using popular surveillance datasets such as KTH \cite{3}, Weisman \cite{4}, IXMAS \cite{5}, UCF-ARG \cite{6}, MHAD \cite{7}, PETS series \cite{8}\cite{9}. These datasets were recorded in controlled conditions, such that an individual actor performs an action in a video clip in nearly an identical manner in terms of camera's position, illumination variation with a simple static background; therefore, the recognition rate is very high (up to 90\%) for almost all methods. 

In the case of classification approaches, videos are classified according to the action a video contains. These approaches are evaluated using newly developed datasets with videos that are either gathered from the web, compiled from YouTube or video clips that were recorded in realistic conditions where illumination conditions are not controlled, using moving cameras and unstated complex background. Datasets belonging to this category include Hollywood \cite{10}, Hollywood2 \cite{11}, UCF sport \cite{12}, UCF50 \cite{13}, UCF101 \cite{14}, HMDB51 \cite{15}, HMDB \cite{16}, and ActivityNet \cite{17}. The diversity of clip content (scale of the actors, objects change positions, etc.), the variation in a camera's motion, and the background obsolete analysis of this category of the dataset are explored. Table 1 summarizes the categories of existing methods and their description related to the utilized datasets.

This paper proposes a video summarization approach based on action recognition where human body actions are first detected and identified. By detecting and recognizing many human body actions  like illustrated in figure 1, the proposed method provides the ability also to summarize these detected action per person, unlike the exixting approach that recognize one action per image. The proposed consists of detecting, recognizing and summarizing  multiple actions made by many persons in the same video. The major contributions of this paper can be summarized as follows:

\begin{itemize}
  \item \relax Preparation of data using an extraction of sequences of each human silhouette in each time .
  \item \relax  Each sequence is exploited to recognize the action using a 3-dimensional CNN model.
\item \relax An action-based video summarization for all persons in the scene
  \item \relax The proposed architecture is employed for video classification(for movies, YouTube videos ) without preprocessing to recognize the action in the video.
\end{itemize}
  The remainder of the paper is organized as follows. The literature overview including sequential-based and CNN-based methods related to our work are presented in section 2. The proposed method is presented in section 3. Experiments performed to validate the proposed method, is discussed in section 4. The conclusion and future works are provided in section 5.   
%

\section{Literature review }

\subsection{Sequential-based methods }

Several methods of human action classification have been proposed. These methods can be classified into three main types: appearance-based, motion-based, and space-time-based methods. The methods that based on montion in the videos  consist of an optical flow compution followed by a motion templates comparison with optical flow results . For the second catrgory, appearance-based methods, the motion history of the images in the sequence is extracted defore comparing it with the active shape models. For space-time methods, the recogntion is made by traning the space-time features to identify the actio. Some researchers have exploited visual search of compact descriptors to recognize the actions; for example, (CDVS) \cite{20} and CDVS feature trajectories of detected human bodies \cite{21}. The global and local features of the image given by CDVS features is a good technique for feature extraction in real-time because of the use of computing optimization. Other authors exploited the saliency of images to select the region that describe an action, followed by a clustering operation classify actions \cite{22}. In another method, the authors propsoed a  non-linear SVM decision tree method that exploited Smith-Waterman and fast HOG3D features of each frame  \cite{23}. Using human motion for action recognition purposes, the authors in \cite{24}, extracted view-invariance between 3D and 2D videos. Then,  3D dense trajectories are extracted from 3D videos and used for recogntion. Based on histogram of oriented gradient (HOG) of motion history image (MHI) and Speeded up Robust Features (SURF) a human action recognition method is proposed in \cite{25}. The authors in \cite{26} proposed an action recognition method using the RGB-D videos as input and the HMM features of the recognition results for summarization step. 

{\renewcommand{\arraystretch}{1.3}%
\begin{table*}[t!]
\caption{{CNN-based methods for action recognition} }

\centering 
\begin{tabular}{c c p{4cm} p{7cm} }
\hline
\multicolumn{1}{p{2.5cm}}{\textbf{Architecture}} & \multicolumn{1}{c}{\textbf{Stream input}}      & \multicolumn{1}{c}{\textbf{Method}} & \multicolumn{1}{c}{\textbf{Input of CNN}}                                        \\ \hline
\parbox[t]{2mm}{\multirow{8}{*}{\rotatebox[origin=c]{90}{2DCNN}}}   & \multicolumn{1}{c}{\multirow{2}{*}{One-Stream}} 
                                                                              & Wang et al. \cite{45} & Saliency-aware \\ \cline{3-4} 
                              &                                                & Akula et al. \cite{49} & Infrared images \\ \cline{3-4} 
                             &                                               & Yang et al.  \cite{40} & RGB \\ \cline{2-4} 
                            
				& \multirow{2}{*}{Multi-Stream}                      
                                                                               &  Simonyan et al.  \cite{41} & RGB image, OF images \\ \cline{3-4} 
                            &                                                  & Wang et al.\cite{46} & RGB  , Optical flow OF ,  MSDI \\ \cline{3-4}
                            &                                                  & Ma et al. \cite{47} & RGB + OF, Human region RGB + OF, Operation + OF \\ \cline{3-4}  
                             &                                               &  Tu et al. \cite{48}  & RGB, Motion RGB \\ \hline

\parbox[t]{2mm}{\multirow{12}{*}{\rotatebox[origin=c]{90}{ 3DCNN}}}   & \multirow{2}{*}{One-Stream}                       

                                                                               &  Ji et al.  \cite{42} & Image sequence \\ \cline{3-4} 
                            &                                                  & Yang et al. \cite{50} & Videos \\ \cline{3-4}
                            &                                                  & Wang et al. \cite{44} & Trajectory map\\ \cline{2-4}

                             & \multirow{2}{*}{Multi-Stream}                      
                                                                        & Sun et al. \cite{39} & RGB images \\ \cline{3-4} 
                                                                             
                            &                                                  & Wang et al. \cite{43} & Motion representation \\ \cline{3-4}
                            &                                                  & Karpathy et al. \cite{37}&Context stream that models low-resolution image,  \newline Fovea stream that processes high-resolution center crop\\ \cline{3-4}  
                             &                                               &   Tejero-de-Pablos et al. \cite{51}& Body joint-based feature (Human Skelton images), \newline Holistic feature (segmented images) \\ \hline

\end{tabular}
\end{table*}\quad

In order to detect and recognize multiple human posture in videos, the authors in\cite{27}  have used local feature descriptors including HOG and Block Orientation. In the same context, authors in \cite{28} extract a new set of the local binary pattern (LBP), histogram of oriented gradient (HOG) and Harlick features. Extracted features are combined to generate another feature using joint entropy-PCA-based method and Euclidean distance. the recognition is made by classifying the selected features using a Multi-class SMV technic. Using Scale Invariant Feature Transform (SIFT) and Histogram for Oriented Gradient (HOG) for extraction of features in addition to Support Vector Machine (SVM) classifier, authors in\cite{29} proposed a method to detect and recognize human actions. In \cite{30}, authors exploit the interest points and the 3D scale-invariant feature transform (3D SIFT) extracted for each point, to classify human action. For recognition of human actions, authors in \cite{31} use a novel representation of videos called high-dimensional vector before applying a dimension reduction operation using compressive sensing with Gaussian mixture random matrix (CS-GMRM). Some authors like in \cite{32} have combined the covariance matrices as local spatiotemporal descriptors and log-Euclidean covariance matrices (LECM) extracted densely from action videos.
In the same context, authors proposed in \cite{33} a behavior recognition approach where spatiotemporal features are extracted using 2D and 3D interest points. In \cite{34}, based on the camera motion estimation, the authors improved dense trajectories in order to recognize actions. Features used in image processing fields \cite{35}, are exploited for video action recognition including 3D histogram of oriented gradient (3DHOG), histogram of optical flow, motion boundary histograms (MBH), 3D-SIFT \cite{33}, and improved dense trajectories (iDT) \cite{34}. The robustness of such features is related to many characteristics. For example, MBH features are useful for camera motion and gives better results that HOF features. Also, iDT has been shown to outperform a combination of HOF and MBH \cite{36}. The effectiveness is also related to the type of videos. For example, previous methods give good results in (UCF-101, HMDB-51) datasets but they can fail in the case of crowded videos and videos with large variations of action categories.

\begin{table*}[t!]
\caption{{CNN-based methods for action recognition} }

\centering 
\begin{tabular}{ p{2cm} p{2cm} p{6cm} p{4cm}}
\hline
\multicolumn{1}{c}{\textbf{Type}}   & \multicolumn{1}{c}{\textbf{Method}} & \multicolumn{1}{c}{\textbf{NNNN}}  &  \multicolumn{1}{c}{\textbf{Dataset}}                                      \\ \hline
\multicolumn{1}{c}{\multirow{3}{*}{Statistical and Texture-based }}    & [13] & LBP-SVM classifier & NUAA and CASIA-SURF  \\  \cline{2-4} 
                                                  & [14] & HOG/ Logistic Regression classifier  & MSU-MFSD  \\ \cline{2-4} 
                                                 & [15] & DOG/Binary classifier & CASIA FASD and MSU MFSD \\  \hline
 \multirow{3}{*}{Motion-based}                      
                 &  [16] & Eye blinking/ Binary classifier  & DLIP  \\  \cline{2-4}
                    & [17] & Lip movement, chin movement and eye blinking/ Average intensity classifier &   ZJU Eyeblink \\ \cline{2-4}
                     & [18] &Facial expression, eye blinking, mouth movement/ 2-way softmax classifier & CASIA-FASD \\   \hline

 \multirow{3}{*}{Frequency-based}                      
                 &  [12] & Time Frequency domain/GMM classifier  & ASVspoof 2019  \\  \cline{2-4}
                    & [19] & 2D DFT/SVM classifier & CASIA, Replay-Attack and MSU \\ \cline{2-4}
                     & [20] &2D FFT/SVM classifier & CASIA-FASD, Replay Attack, 3DMAD and UVAD \\   \hline

 \multirow{3}{*}{Color-based}                      
                 &  [21] & color texture Markov feature/SVM classifier  & CASIA, MSU MFSD, Replay-Attack and OULU-NPU  \\  \cline{2-4}
                    & [22] & Color texture/ Softmax classifier &  CASIA, Replay-Attack and MSU MFSD \\ \cline{2-4}
                     & [23] &  &  \\   \hline

 \multirow{3}{*}{Color-based}                      
                 &  [24] & Context descriptor-SVM classifier & self-collected anti-spoofing datasets  \\  \cline{2-4}
                    & [25] & Context and Texture Information-SVM classifier &  CASIA, Replay-Attack \\ \cline{2-4}
                     & [26] &  Local appearance and context descriptors/  nonlinear SVM classifier & NUAA, CASIA, Replay-attack, and binocular camera \\   \hline

 \multirow{3}{*}{Color-based}                      
                 &  [27] & LSTM-SVM  & CASIA and replay attack \\  \cline{2-4}
                    & [28] & LBP-TOP-SVM &  Replay-attack and MSU \\ \cline{2-4}
                     & [29] & High level semantic feature/simple classifier & OULU-NPU, CASIA-SURF and SiW \\   \hline

\end{tabular}
\end{table*}\quad

\subsection{Convolutional-neural-network-based methods} 

To improve the action recognition performance, recent works have employed various deep learning models \cite{37}, \cite{38}. Since human actions are extracted from multiple movements of human body or some parts of it, it is necessary that the recognition process should involve video browsing over time to learn the patterns of the visual appearance changes \cite{39}. To achieve this, existing deep learning models based on 2D convolutional networks can be extended into 3D domain to capture the temporal information \cite{40}. For example, the authors in \cite{41} use the motion in consecutives frames to extract information for action recognition using a two-stream ConvNet architecture that incorporates both spatial and temporal networks.

In the same context, authors in \cite{42} proposed a 3D CNN{\textendash}based method using stacked frame cubes of actions to capture 3D spatial-temporal action signals. In addition, Wang et al. \cite{43} built up a temporal pyramid pooling based CNNs for action representation comprising of an encoding, a pyramid pooling and a concatenation layer, which can transform both motion and appearance features. Moreover, in \cite{44}\textbf{,} trajectory-pooled CNNs was designed by fusing Improved Trajectories into CNNs architecture, while an adaptive recurrent convolutional hybrid networks was proposed consisting of a data module and a learning module.

\bgroup
\begin{figure*}[t!]
    \centering
 \includegraphics[width=0.75\textwidth]{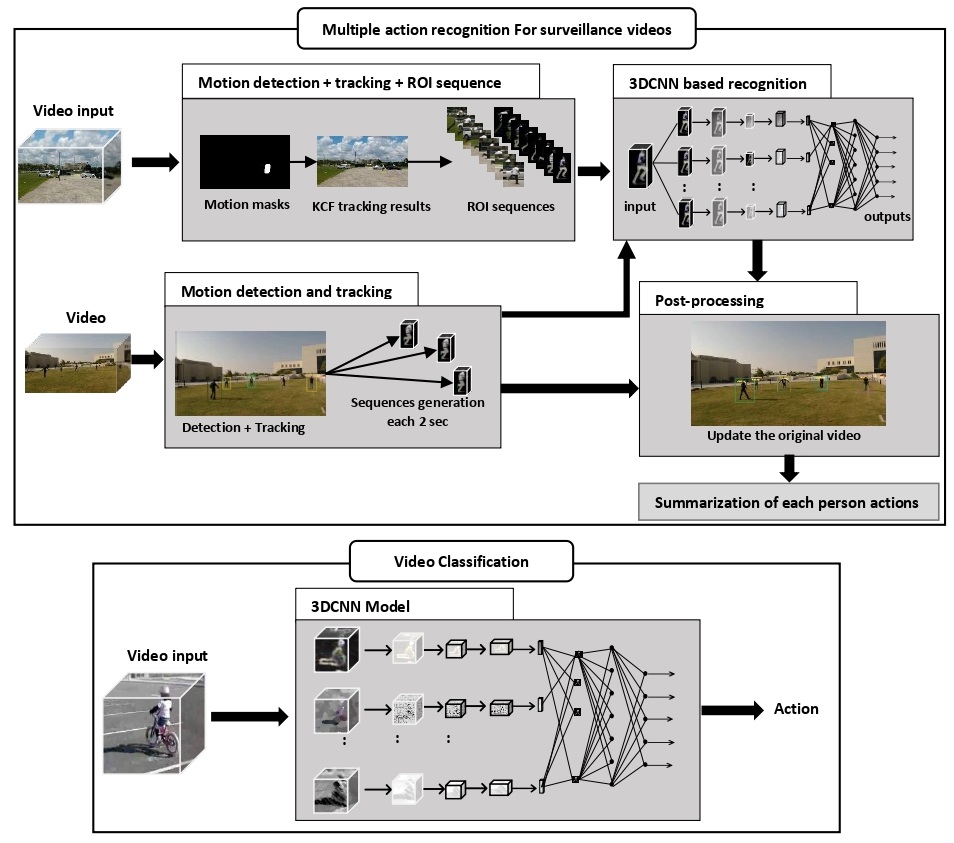}
\caption{{Diagram of the proposed multi human action detection and recognition method.}}
\end{figure*}
\egroup

For the same reason, many methods use several input videos for deep learning modelling in order to recognize the human action. In \cite{45}, Wang el al. use a combination of saliency-aware, 3D CNN model and long short-term memory (LSTM) to recognize human action. Another example of many features used with CNN model was proposed in \cite{46} where the authors exploit the original frames, optical flow, and motion stacked difference mage (MSDI) as inputs of CNN model for human action recognition. Ma et al. \cite{47} propose CNN-based methods for action recognition that use six-stream features in order to use a general action recognition where many inputs are used including the full images, human image representing just the human body, operation region represent the part of body in action, and the optical flow result of each one of the previous features. In the same category of methods that use multi-stream CNN-based technic, Tu et al. \cite{48}propose a multi-stream convolutional neural network architecture to recognize human action. The proposed approach detects human body in action and the region of interest corresponding to the moving parts. The fusion of all CNN results is used to recognize the human action. Other methods use infrared images to recognize the human action \cite{49}. In order to summarize the sport player actions, authors in\textbf{\space }\cite{51} use a 3D CNN model with two streams as input the first represented the video of human body skeleton and the second represent the video of segmented data\textbf{.} For summarizing all the categories of human action recognition methods based on convolutional neural networks, Table 2 illustrates each category and the features used as well as the stream used.

\begin{figure}[h]
\centering 
 \includegraphics[width=0.4\textwidth]{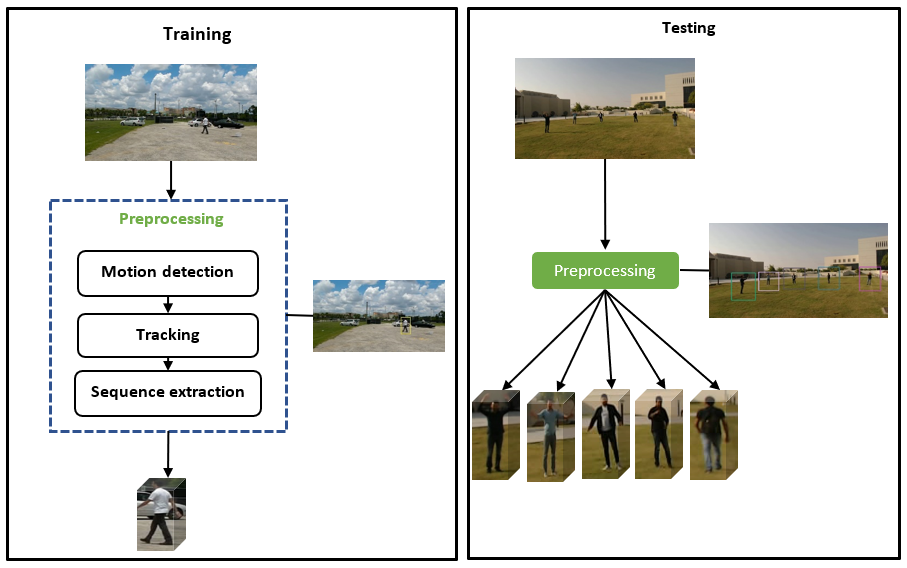}
\makeatother 
\caption{{Sequence extraction process.}}
\label{f-bce59cf1920b}
\end{figure}
 
\section{Proposed approach}
 This section introduces the proposed approach which consists of four phases as illustrated in figure 2. In the first stage (i.e., the preprocessing phase), we extract detected and tracked persons to generate target-region sequences from the videos. Then, the proposed convolutional neural network architecture which includes both 3-dimensional and 2-dimensional networks where the 3DCNN is used to recognize action from the generated sequence videos of each person while the 2D counterpart is used to recognize the actions based on the Motion History Images (MHIs) of the detected action(s). The third phase, referred to as the post-processing stage,  is a presentation of a multi-action detection and recognition method for many targets in a video.\textbf{\space }In the last section (action classification), the proposed approach uses our proposed 3DCNN without any pre-processing on videos collected from the web and YouTube especially to deal with the more complex video contents.

After the recognition stage, the summarization of actions from the scene is performed for each person. The video summarization is based on the recognized action of each person at each time in the video. For that, this summarization can be represented by frames(each frame illustrates an action made) or just the timeline of all actions made by each person during his presence in the scene. 

\subsection{Pre-processing stage } In this work, we introduce a new representation of data to be suitable for multiple recognition of human action(s). The proposed representation is tailored towards surveillance videos captured by a fixed camera installed in private or public scenes with moving objects targeting surveillance video to recognize human actions. In addition, we are interested in the human motion regions because the human body can occupy a small and specific region in the video while the existing approaches use the entire video sequences to recognize human actions. 

In order to prepare data for convolutional neural networks for training, we isolated human bodies in the video to extract a sequence or clip of body motions during their action and their presence in the video. The detection is made using the method described in \cite{18,181} which consists of background modeling and subtraction before segmenting the moving objects. As a result, we obtained the region of moving objects as presented in figure 2 and figure 3. The extracted sequence of the detected person is the input of the proposed neural network architecture.

\begin{figure*}[t!]
\centering 
 \includegraphics[width=0.8\textwidth]{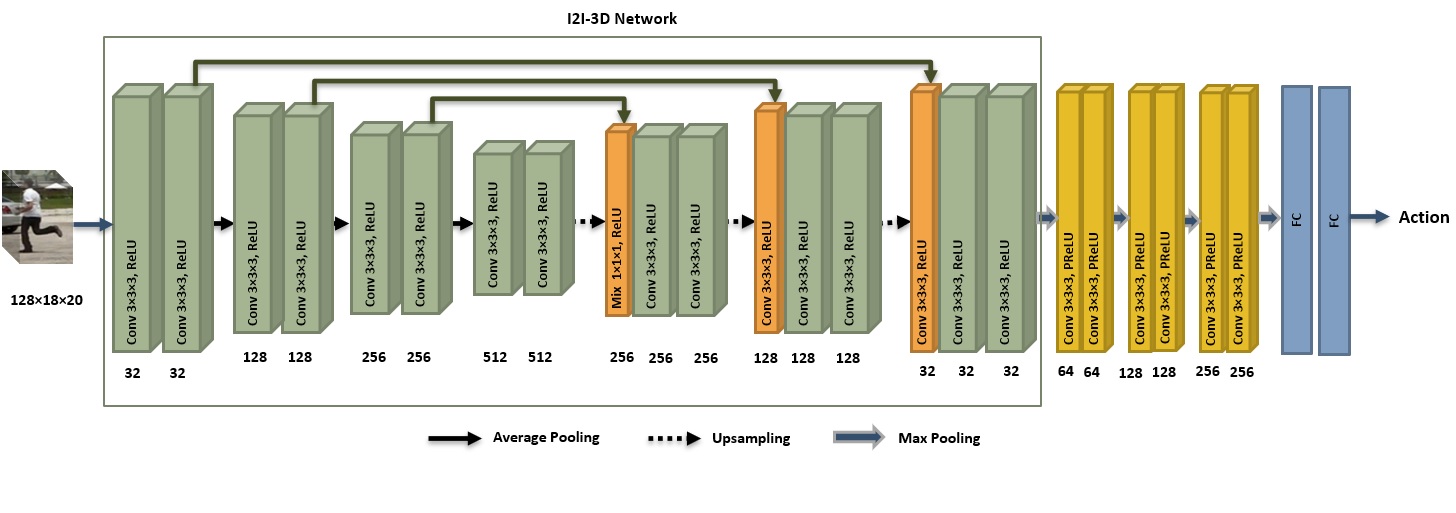}
\caption{{The proposed 3DCNN architecture.}}
\label{f-5145e7cdeba2}
\end{figure*}

\subsection{3DCNN Model}
Depending on the applications, the selection of the optimal CNN architecture is challenging. The proposed deep learning-based approach involves the preprocessing of action videos before feeding them to the convolution neural network. The preprocessing consists of extracting the target region that contains human bodies in action, followed by resizing the data before creating NumPy video. A 3-Dimensional Convolution Neural Network (3DCNN), which is supervised learning with a multistage deep learning network, has been implemented. 3DCNN is capable to learn multiple invariant features from the input's videos.
 
The proposed 3DCNN-based architecture illustrated in Figure 4 consists of two parts, a 3D encoder-decoder network followed by the classification block of convolutional-max-pooling layers. 

The 3D encoder-decoder network, inspired by \cite{56}, is based on VGG16 architecture for generating features and increasing a greater spatial extent. The encoder-decoder allows complex multiscale learning from a 3D volume that represents the video. Also, the skip connections allow learning with high-resolution features and generate a new higher resolution response with multi-scale influences. Also, the mixing layers combine the output of two blocks and minimize the multiscale objective function.

The second part of the proposed model of classification block, as illustrated in figure 4, is composed of three 3D convolution-pooling units, six convolutional layers and three MaxPooling layers, one flattened layer, and two fully connected layers. The output layers consist of ten neurons that represent the number of actions. This block allows a transaction from 3D representation to flatten data which makes the model learning from multiscale layers also using a complex architecture that performant learning.

For the last six convolutional layers, we have used a parametric Rectified Linear Unit (PReLU) as the activation function, which is a generalized parametric formulation of ReLU. This activation function adaptively learns the parameters of rectifiers and improves the accuracy at negligible extra computational cost. Only positive values are fed to the ReLU activation function while all negative values are set to zero. PReLU assumes that a penalty for negative values, and it should be parametric. The PReLU function can be defined as:  
\begin{eqnarray*}f(y_i)=\left\{\begin{array}{l}y_i\;if\;y_i>0\;\\a_iy_i\;if\;y_i\leq0\end{array}\right.\end{eqnarray*}
where $a_i $ controlling the slope of the negative part. When  $a_i $= 0, it operates as an ReLU; when $a_i $ is a learnable parameter, it is referred to as Parametric ReLU (PReLU). Figure 5 shows the shape of PReLU activation. If ai is a small fixed value, PReLU becomes LReLU ($a_i $= 0.01). PReLU can be trained using the backpropagation concept.

For the training and testing, we have used the preprocessed data from KTH, Weizmann, and UCF-ARG datasets. The model is trained using \textit{CrossEntrpy} with a batch size of 64 examples, a learning rate of 0.001 like illustrated in Table 3.

\subsection{Multi-Action detection and recognition and post-processing.} In the case of many moving objects or persons in the scene, we detect and track each person in the scene to generate a sequence of each one of them. The tracking helps generate a sequence of the same detected person during his presence in the scene. To achieve this, we have used the extended version of kernelized correlation filter (KCF) based method for tracking \cite{19}. From the tracking results, we are able to extract and generate the sequence of each person which represents the human body actions. Recognizing the activity of different persons in one scene has the added advantage of extracting a sequence for each person.

The proposed approach using the proposed pre-processing can be applied for the recognition of each of these actions. The detection and multiple tracking allow recognizing many actions made by many persons at the same time.

\begin{figure}[h!]
\centering
 \includegraphics[width=0.5\textwidth]{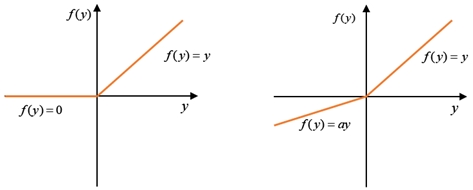}
\makeatother 
\caption{{Activation functions of ReLU and PReLU. (Left) ReLU. (right) PReLU. In the case of PReLU, the coefficient a is learned from the data.}}
\end{figure}

After the extraction of each person sequence, the recognition is performed using the proposed 3DCNN architecture using the sequence as input. Each RGB video (or sequence) may contain some redundant (non-useful) information like the static background. For that, we apply the 3DCNN also on the RGB sequence without background called BS-video. Also, in order to minimize the time of recognition we use Motion History Images (MHI) extracted from BS-video and apply the 2DCNN version of the proposed architecture. The use of MHI improves the recognition rate. Figure 6 illustrates the use of different inputs of the deep learning model.

\subsection{Action classification.}The main difference between surveillance videos and other types of videos from movies, phone cameras, etc. is related to the stability of the camera as well as the coverage scope. Surveillance video focuses on places that can be widely covered. The other category of videos collected from the Web and YouTube or movies usually consist of crowded videos and videos with large variations of action categories. For these kinds of videos, the content is very difficult to be analyzed, making the extraction of patterns and the recognition of the action extremely very difficult. Various researchers have attempted to classify these videos based on the action contained in the videos. Several datasets including YouTube, UCF101, HMDB51 contain millions of videos that have been constructed.

This paper proposes to recognize the action(s) from these videos using our proposed 3DCNN model. The videos are introduced to the neural network without any preprocessing and the recognition is made by learning the patterns using all the datasets. From the experimental results, the proposed model provides an improved recognition rate compared to the state-of-the-art-methods.

\begin{figure}[h]
\centering 
\includegraphics[width=0.45\textwidth]{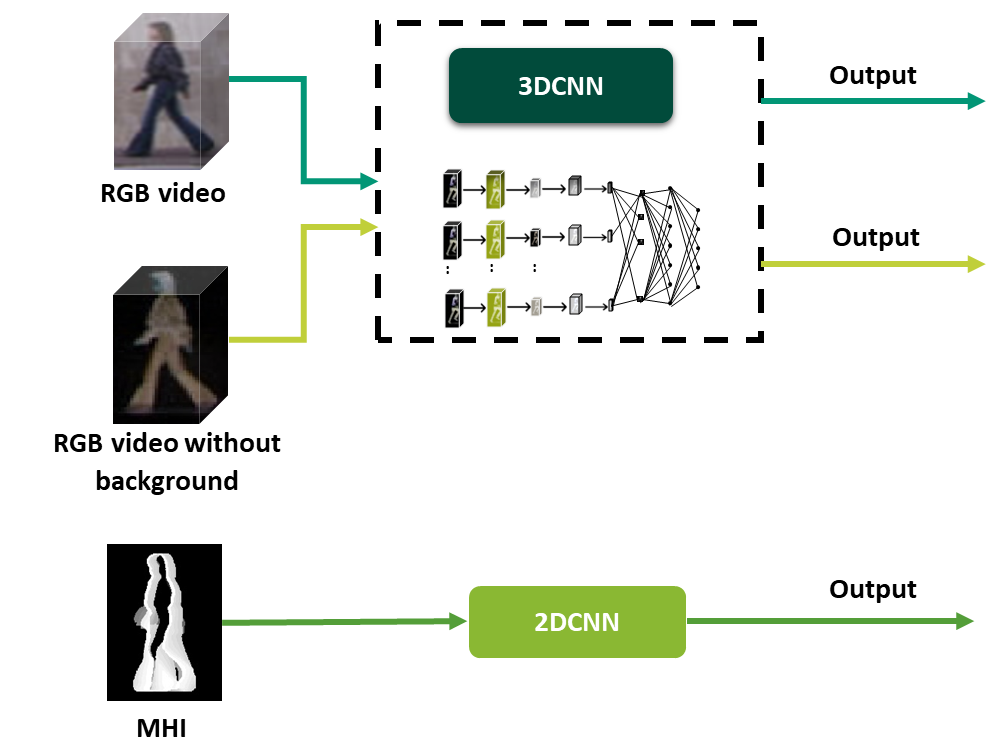}
\makeatother 
\caption{{CNN models with different inputs}}
\label{f-a229289d135b}
\end{figure}

{\renewcommand{\arraystretch}{1.3}%
\begin{table}[t!]
\caption{{Hyperparametres of the proposed model} }
\centering 
\begin{tabulary}{\linewidth}{p{\dimexpr.27\linewidth-2\tabcolsep}p{\dimexpr.15\linewidth-2\tabcolsep}p{\dimexpr.15\linewidth-2\tabcolsep}p{\dimexpr.12\linewidth-2\tabcolsep}p{\dimexpr.2\linewidth-2\tabcolsep}}
\hline
Method & Layers & Optimizer & LR & Batch size\\ \\
\hline 
3DCNN model &  35 &   Adam &   0.001 &   64\\
\hline
\end{tabulary}
\end{table} \quad

\subsection{Video summarization based on recognized actions}
Video summarization aims to generate a short version of a video as a representation, using key frames of important subsequences \cite{51,52}. This summarization provides a rapid view of the information contained in a large video. It also provides a good evaluation for users of the video and provides knowledge regarding the topic and the most important content in the video\cite{53,54}. Considering the information contained in each video, many methods have been developed using several techniques\cite{55}.

Summarizing videos using human actions is propsoed in this paper. After recognizing each action we summarize this action using a timeline. These operation is performed for each person action during his presence in the scene.
For the surveillance video that contains persons acting in the scene tha sumamrization is a representation of the actions made by each one of them.
For the video classification (e.g. recognition of action within the video without analyzing the content), the summarization is performed by representing, using a timeline, the actions in each shot(subsequence) of the video.

\section{Experimental results }
This section presents the evaluation of the proposed action recognition approach. The experimentation was carried out using well known KTH, Weizmann, UCF-ARG, Interaction, IXMAS, and MHAD datasets. In addition, the 3DCNN model has been trained and tested on UCF101, Hollywood2, Youtube, and HMDB51 datasets. The 3DCNN model, with the new representation, was then evaluated in terms of the detection and recognition performances of multiple person actions at the same time. Further, the accuracy of the proposed method was evaluated using three different inputs such as RGB video, background subtraction video, and MHI features for surveillance videos. The results of the proposed method were then compared with the state-of-the-art methods that are based on surveillance videos, as well as methods based on action video classification. 

For the surveillance videos, the obtained accuracies for each dataset has been compared with a set of recent works. On KTH dataset the obtained results are compared with the methods proposed in \cite{27}, \cite{28}, \cite{30}, \cite{21}, \cite{32}, \cite{23}, \cite{60}, \cite{62}, \cite{64}, \cite{58}, \cite{59}, \cite{61}, \cite{63}, and \cite{65}. For Weizmann dataset we compare the results with the methods in \cite{31}, \cite{25}, \cite{28}, \cite{59}, \cite{64}, \cite{58}, and \cite{65}. Also, for UCF-ARG, the obtained results are compared with the methods in \cite{29}, \cite{65}, and \cite{66}. We compared our method with the method in \cite{67}, \cite{68}, and \cite{65} for Interaction dataset. For IXMAS dataset, the proposed method has been compared with \cite{69}, \cite{63}, \cite{65}, \cite{70}, \cite{71}, \cite{72}, and \cite{73}.

The non-surveillance videos dataset which are for video action classification, we compared our approach with different methods on four datasets including UCF101, Hollywood2, YouTube and HMDB51. Some methods are trained and tested on all the four datasets while some others are trained and tested on one or two of these datasets. The list of methods that we compare with for video action classification are : \cite{37}, \cite{39}, \cite{40}, \cite{41}, \cite{43}, \cite{44}, \cite{45}, \cite{46}, \cite{47}, \cite{48}, \cite{50}, \cite{61}, \cite{62}, \cite{63}, \cite{64}, \cite{65}, \cite{74}, and \cite{75}.

\bgroup
\begin{figure}[t!]
\centering
 \includegraphics[width=0.45\textwidth]{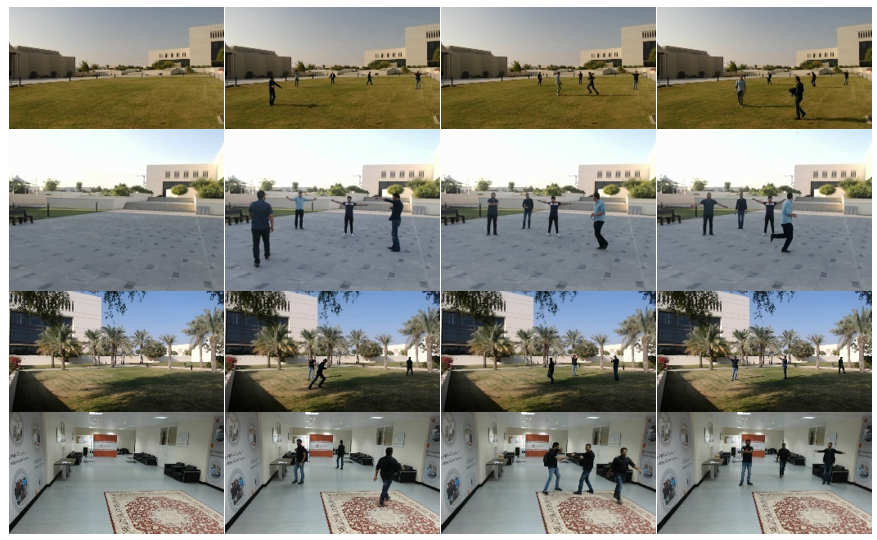}
\makeatother 
\caption{{MHAD dataset \cite{7}. first row: background images, rest rows: frames from the dataset.}}
\label{f-e0b68b687d2b}
\end{figure}
\egroup

\begin{figure*}[t!]
    \centering
    \subfloat[KTH]
        {
        \includegraphics[width=12cm,height=16cm,keepaspectratio]{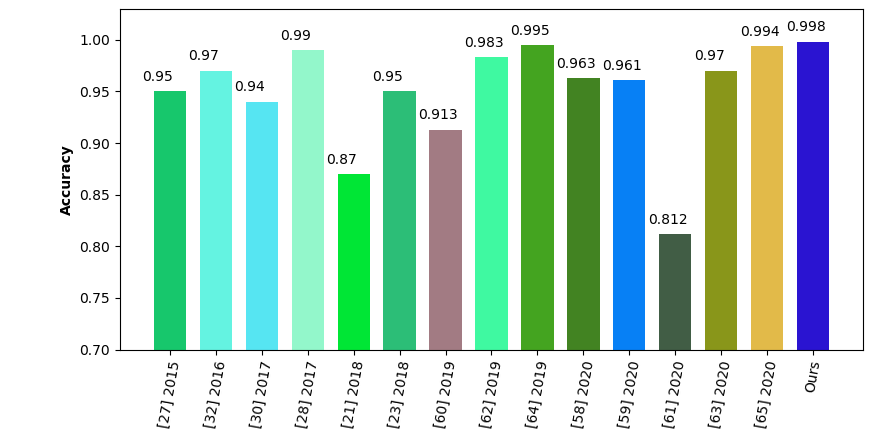}
        \label{fig}
        }\\
    \subfloat[Weizmann]
        {
        \includegraphics[width=8cm,height=3cm]{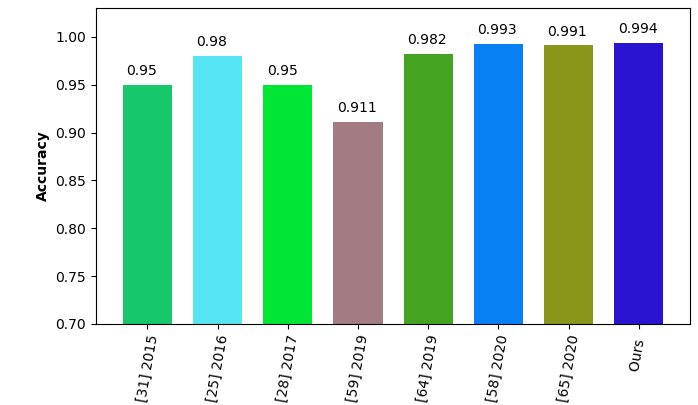}
        
        } 
    \subfloat[UCF-ARG]
        {
        \includegraphics[width=4cm,height=3cm]{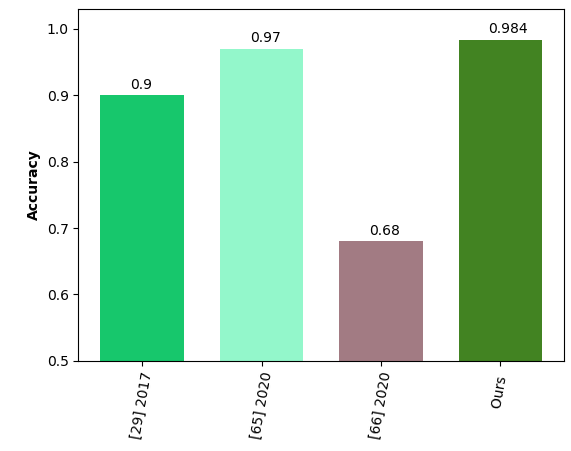}
        
        }\\
    \subfloat[IXMAS]
        {
        \includegraphics[width=8cm,height=3cm]{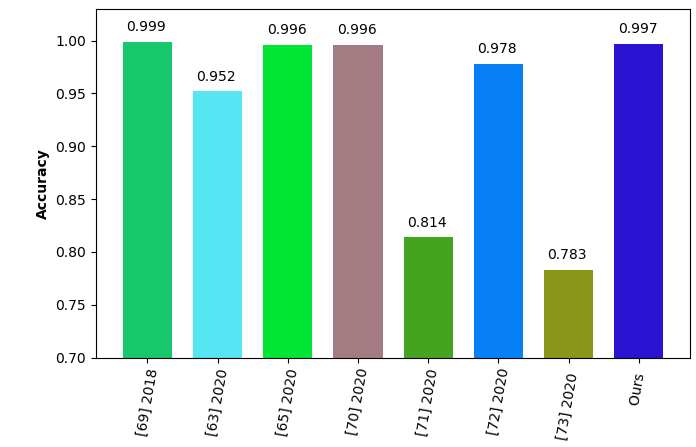}
        \label{fig}
        }
    \subfloat[Interation]
        {
        \includegraphics[width=4cm,height=3cm]{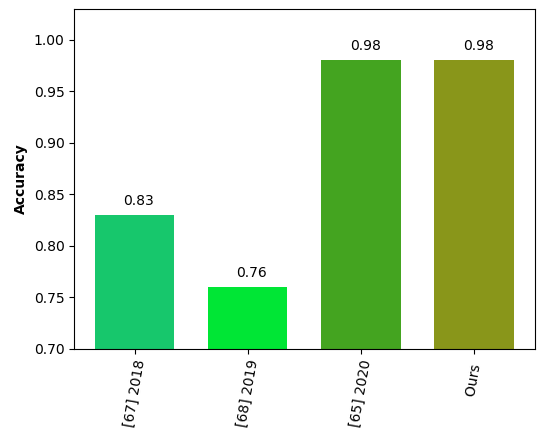}
        \label{fig}
        }
    \caption{Recognition performance on each Dataset with comparison with state-of-the-art methods.}
    \label{fig:globfig}
\end{figure*}

\subsection{System setup}
To recognize human actions, model only regions of interest from each video, in the three datasets KTH, Weizmann, and UCF-ARG were extracted. This provides a new representation of data for the deep learning model. Each video was divided into shorts clips of 1 to 2 seconds that can represent the time of action, then converted to an\textit{ NPZ} file. The number videos for each action, with the new representation is from 200 to 300 for each action.

The proposed system has been implemented with python programming language using a laptop with GPU NVIDIA 1070 GTX. The training has been made using the new representation on surveillance videos while  MHAD dataset has been used for testing multiple human action detection and recognition. Figure 7 represents some frames from each video of MHAD dataset with also the background. For the training phase, we use two splits 80\% and 90\% of data where 10\% and 5\%of data is for validation and 10\% and 5\% for testing. Using 100 epochs, the accuracy of the proposed 3DCNN-based approach reaches a good accuracy using different inputs respectively.

To ensure good detection and recognition performance, the new presentation of the datasets was used by dividing data between training, validation, and testing. Also, to represent each action in many different situations, a set of videos with individual actions made by different actors, was used.

The recognition using surveillance videos has been performed using the detection of persons in the scene then recognize the action of each one. While the recognition of action in the other types of videos like UCF101 has been made by classifying the action in it using the 3DCNN proposed model.

The summarization of each video depending on the type of video, for surveillance videos we summarize the people's actions during their presence in the scene, but for videos like movies, the summarization is the description of the action in each shot of the video.

\subsection{Evalutation on surveillance data }

\subsubsection{Single human action recognition}

In order to evaluate the proposed 3DCNN model on the surveillance videos, the training exploits different existing datasets for good learning of human actions. The video surveillance datasets used include KTH, Weizmann, and UCF-ARG, IXMAS, and Interaction.

The comparison of the proposed method has been performed also on each dataset, and the obtained accuracies are presented in Figure 8. From the presented figure, it can be observed that many methods succeed to recognize the actions with a good performance for all datasets. For the KTH dataset, which is a simple dataset that contains simple actions captured from the same angle of view, the methods [28], [64], [65], as well as the proposed method, reach better recognition rates. For instance, the Weizman dataset shows that all methods achieve good performance where the accuracy rate exceed 90\% for all methods, also reach 99\% for [85], [65], and the proposed method. The high accuracies for the two datasets come from the fact that the two datasets contain videos with simple backgrounds.

For the remaining dataset including Interaction and IXMAX that contains the action of interaction between more than one person, the accuracies depend on the methods used for interpreting the action in these types of videos. For example, in the IXMAS dataset, the proposed method as well as the methods in [69], [65], and [70] succeed to recognize actions with high accuracies. Where the Interaction dataset can be considered more complex than IXMAS which explains the accuracies in [67] and [68]. While the proposed method and the method in [65] achieve more improved results. 

{\renewcommand{\arraystretch}{1.3}%
\begin{table}[h]
\caption{{Accuracies results using different inputs features On MHAD dataset} }
\centering 
\begin{tabulary}{\linewidth}{m{29mm}m{19mm}m{25mm}}
\hline
  \textbf{Training setting} &  \textbf{Accuracy (\%)} &  \textbf{Accuracy  (\%) with data augmentation}\\
\hline
3DCNN (RGB video) &   87.71\% &   93.87\%\\\cline{1-1}\cline{2-2}\cline{3-3} 
3DCNN (background subtraction video) &  96.22\% &   98.48\%\\\cline{1-1}\cline{2-2}\cline{3-3} 
CNN (Motion History Image) &   96.36\% &   99.14\%\\\cline{1-1}\cline{2-2}\cline{3-3}
\end{tabulary} 
\end{table}\quad
\normalsize

\small
\begin{figure}[!t]
\centering 
 \includegraphics[width=0.48\textwidth]{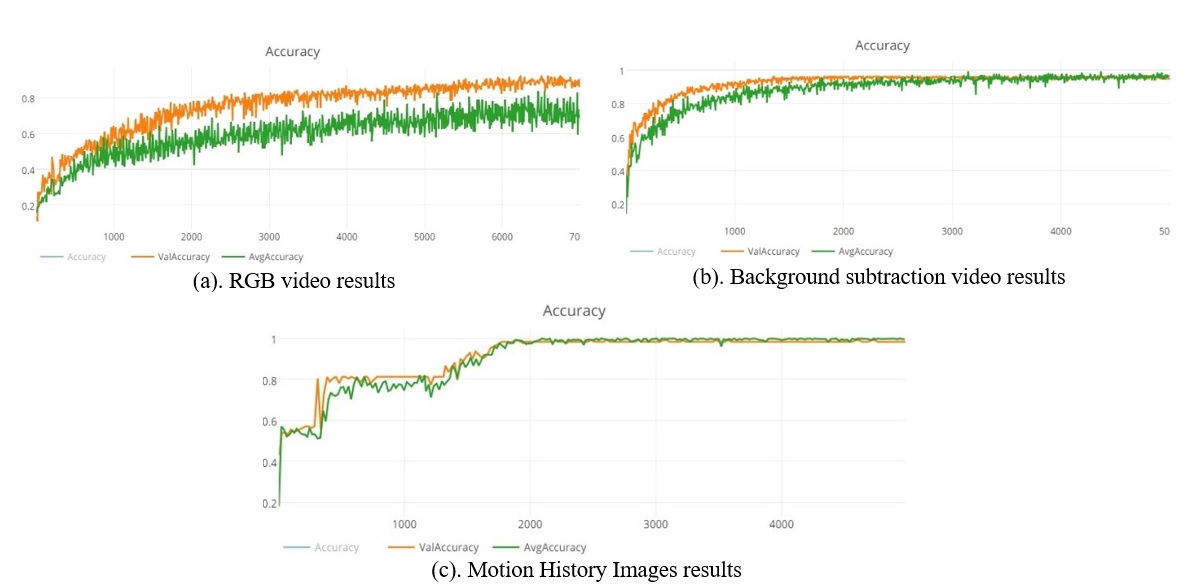}
\makeatother 
\caption{{Accuracies of the proposed model using different inputs}}
\label{f-0ca241bd7c83}
\end{figure}
\normalsize

\subsubsection{Multiple human action recognition }

For testing the accuracy of the multiple human action recognition method, the MHAD dataset has been used while the KTH, Weizmann, and UCF-ARG datasets, with the new representation, are used for training. The dataset contains 3 to 5 persons acting in the scene. In addition, three of the videos are taken in an outdoor environment with one in an indoor setup. 

\small
\begin{figure}[h]
\centering 
 \includegraphics[width=0.48\textwidth]{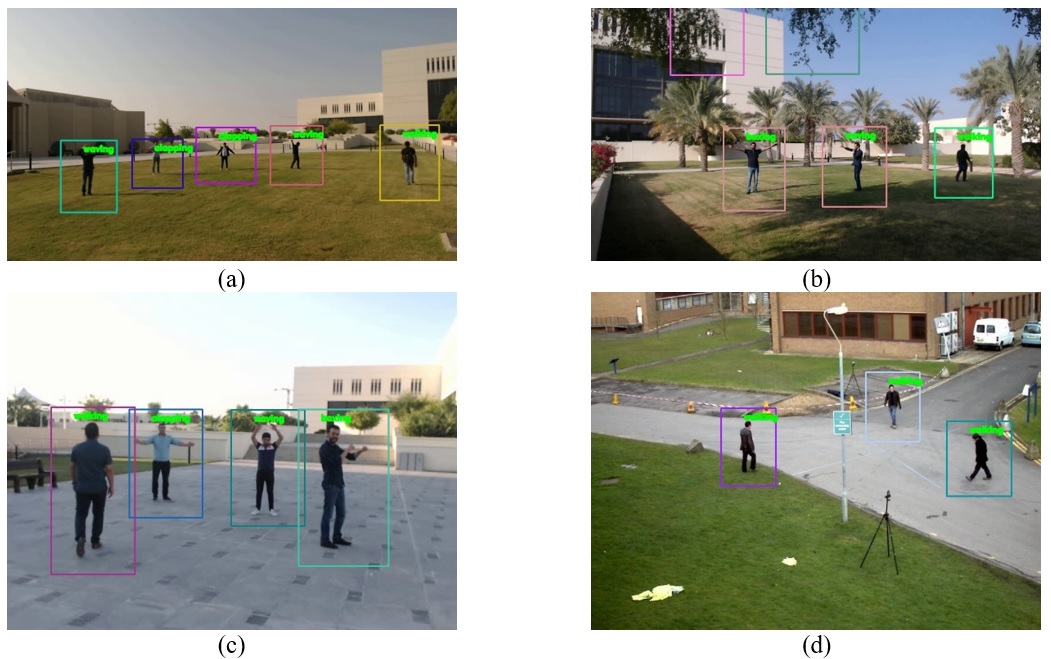}
\makeatother 
\caption{{Detection and recognition results from MHAD \cite{7} and PETS \cite{9} datasets.The block represents the detected human action and the corresponding action.(a), (b)and (c) videos from MHAD dataset. (d) video from PETS 2009 dataset.}}
\label{f-e8f580287c5b}
\end{figure}
\normalsize

Based on different features used as inputs for our 3DCNN and 2DCNN models, the accuracy increases when using the simpler features (sequence of motion of each person without a background as the input of 3DCNN model and motion history images of each person sequence as the input of 2DCNN model) with our representation of data. The RGB video and background subtraction video were used to evaluate the robustness of the proposed approach. For that, the results shown in Table 4 show the accuracy of human action recognition using the proposed model. From the table, it can be clearly observed that the accuracy increases from 87\% to 96\% when the background is subtracted from the RGB video and use the results(video without background)  as input. However, the background might contain non-useful information, therefore, remove the background increases the accuracy. With the 2D version of our CNN model, the Motion History Image was used, represented a summarization of the motion in the video, as input for the CNN model. The performance accuracy is better than the background subtraction counterpart; however, depending on the information (the content of the image) in each MHI that can be different from an action to another. For example, the motion history of the action of walking where the all-body move is different from the action of waving two hands was just the two hands are moving. The accuracy increases also when we augment the number of training data, as we can see the accuracy improved by 6\% for (3DCNN- RGB video) and by 4\% for 2DCNN+MHI. 

\begin{figure*}[!t]
\centering 
 \includegraphics[width=0.6\textwidth]{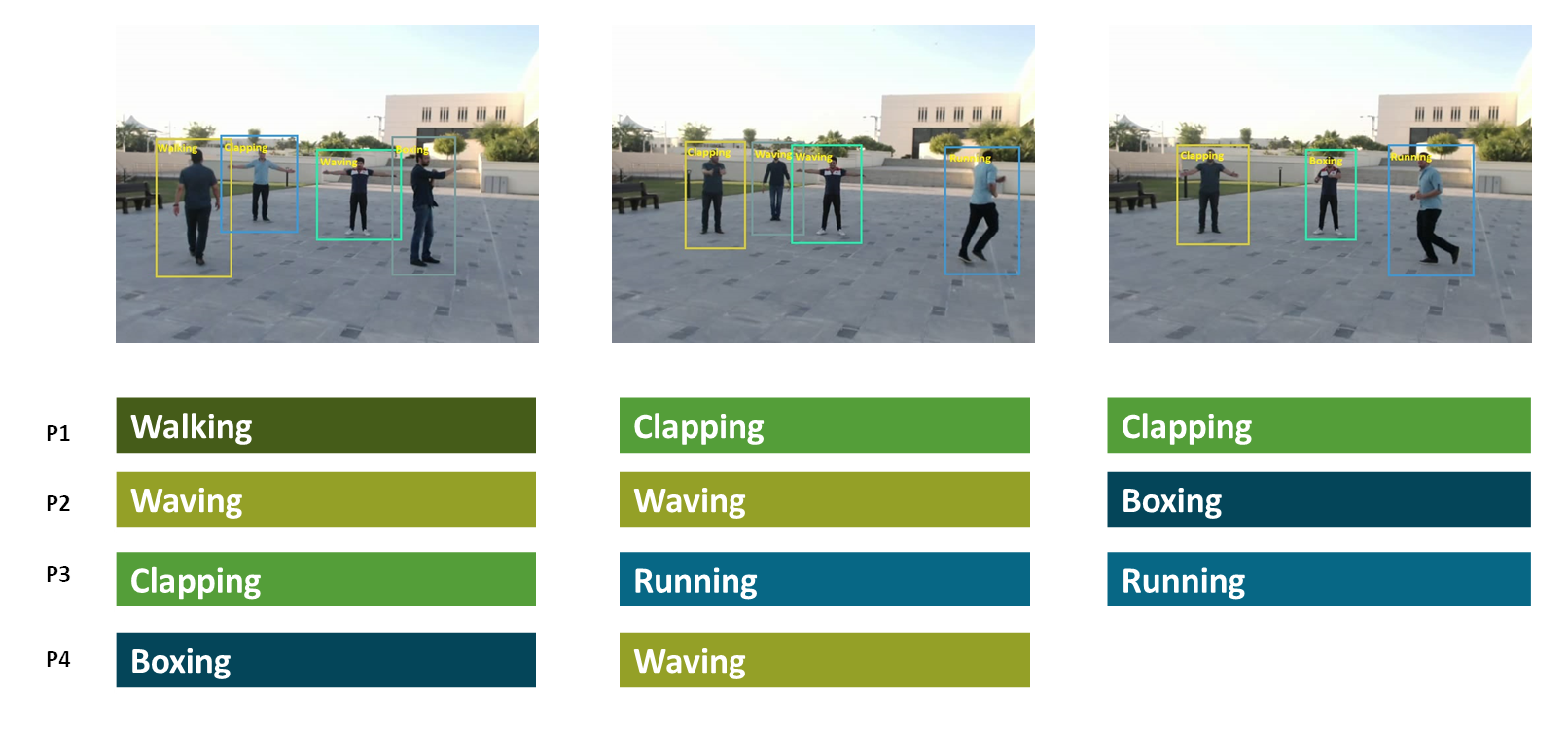}
\makeatother 
\caption{{Summarization of the detected actions made by each person during his presence in the scene.}}
\label{f-0ca241bd7c83}
\end{figure*}

Figure 9 shows the training and validation accuracies graphs using the proposed approach for different inputs including RGB video, Background subtraction video, and MHI when the average recognition rates are 88\%, 96\%, and 99\%, respectively. By removing the background from the RGB video, the accuracy is increased.

The 3DCNN model allows the system to recognize the action of each one during his presence in the scene. Figure 10 shows some examples of detected persons and their actions. Tests are made on three videos from the MHAD dataset and one video from the PETS2009 dataset. The visualized results and accuracy results, shown in Figure 10 and Table 4, reveal that the proposed approach detects and recognizes multiple human actions with an attractive accuracy. In addition, the proposed approach can be improved to be used to recognize multiple human actions in real-time. 

As represented in Figure 11, in the testing part a sequence of each detected person is generated and the action is recognized using the proposed model. The succession of analysis (motion detection, motion tracking, and the proposed architecture) provides a multi human action recognition. After that, each action can be represented in the original video. Also, in order to summarize the detected and recognized actions, each action is saved during the entire time of the video. Figure 11 represents a summarization using graphs of the
recognized actions of each person during his presence in the scene.
\begin{figure}[h]
\centering \makeatletter\IfFileExists{images/y.jpg}{\includegraphics[scale=0.35]{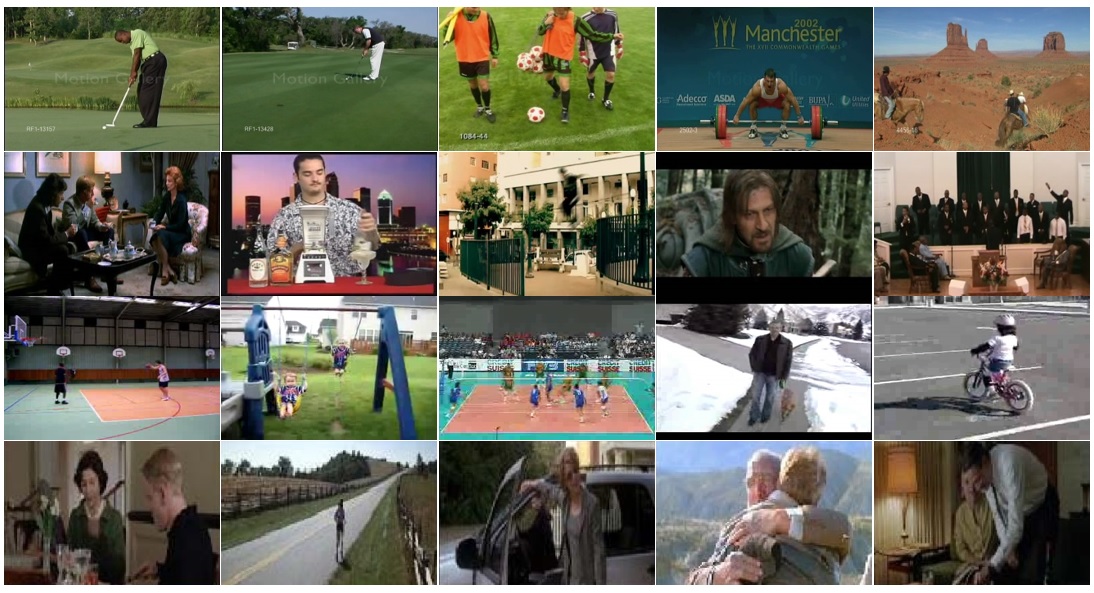}}{}
\makeatother 
\caption{{Examples of human actions used in the experiments. Rows from top to bottom are examples of UCF101, HMDB51, Youtube, and  Hollywoood2 datasets, respectively.}}
\label{f-04784e581486}
\end{figure}

\subsection{Evaluation of action recogntion using video classfication  }

Because of the complexity of the Movies or YouTube videos captured by moving cameras, the variation of points of view, the illumination changes, and video produced by a jitter camera. Many approaches use the entire video to classify the action without analyzing the content of the videos. This work proposes to deploy our 3DCNN model to recognize the actions in these videos. To validate and evaluate our proposed method for video action classification we exploit the other datasets collected from Movie and YouTube and use the video as there are without any preprocessing (no analysis of the content of the videos). We do training and testing on UCF101, UCF sport, YouTube, and HMDB51 datasets. Figure 12 illustrates some example from the used datasets. The Obtained results were compared with the state-of-art methods (video action classification approaches) shown in Table 5. From the accuracy results in table 5, we can observe that the proposed methods can recognize action even without preprocessing. Also, the recognition rates reach 98\%, 91\%, and 97\% for UCF101, Hollywood, and YouTube Datasets respectively.

For summarization of this type of video, we collect a set of scenes from the datasets used and we recognize the action in each subsequence, after that, we represent the action recognized in each time in the video like illustrated in figure 13.

\begin{figure}[h]
\centering \makeatletter\IfFileExists{images/summa.jpg}{\includegraphics[scale=0.5]{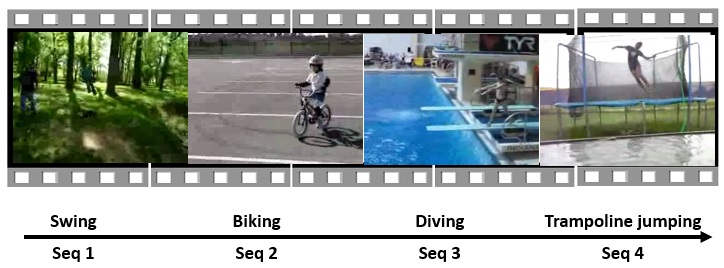}}{}
\makeatother 
\caption{{Action-based video summarization results on videos contains diffrent scenes .}}
\end{figure}

{\renewcommand{\arraystretch}{1.3}%
\begin{table*}[t!]
\caption{Action recognition results on UCF101, Hollywood2, HMDB51 and YouTube datasets comparing with the state-of-the-art methods}
\centering 
\begin{tabulary}{\linewidth}{p{\dimexpr.25\linewidth-2\tabcolsep}p{\dimexpr.12\linewidth-2\tabcolsep}p{\dimexpr.12\linewidth-2\tabcolsep}p{\dimexpr.12\linewidth-2\tabcolsep}p{\dimexpr.12\linewidth-2\tabcolsep}}
\hline
 \multicolumn{1}{l}{\multirow{2}{\dimexpr(.25\linewidth-2\tabcolsep)}{\textbf{Method}}} & \multicolumn{4}{p{\dimexpr(.3\linewidth-2\tabcolsep)}}{\textbf{ Acuraccy(\%)}}\\\cline{2-5} & \textbf{UCF101} & \textbf{Hollywood2} & \textbf{HMDB51} & \textbf{YouTube}\\\cline{1-1}\cline{2-2}\cline{3-3}\cline{4-4}\cline{5-5}

 Karpathy et al. (2014) \cite{37} &  65.4\% &  - &  - &   -\\\cline{1-1}\cline{2-2}\cline{3-3}\cline{4-4}\cline{5-5}
Sun et al. (2015) \cite{39} &  88.1\% &  - &  59.1\% &  -\\\cline{1-1}\cline{2-2}\cline{3-3}\cline{4-4}\cline{5-5}
Yang et al. (2017) \cite{40} &   92.0\% &  - &  64.5\% &  -\\\cline{1-1}\cline{2-2}\cline{3-3}\cline{4-4}\cline{5-5}
Simonyan et al. (2014) \cite{41} &   88\% &  - &  59.4\% &  -\\\cline{1-1}\cline{2-2}\cline{3-3}\cline{4-4}\cline{5-5}
Wang et al. (2106) \cite{43} &  - &  67.5\% &  59.7\% &  -\\\cline{1-1}\cline{2-2}\cline{3-3}\cline{4-4}\cline{5-5}
Wang wt al. (2015) \cite{44} &  91.5\% &  - &  65.9\% &  -\\\cline{1-1}\cline{2-2}\cline{3-3}\cline{4-4}\cline{5-5}
Wang et al (2017) \cite{45} &  84.0\% &  - &  55.1\% &  -\\\cline{1-1}\cline{2-2}\cline{3-3}\cline{4-4}\cline{5-5}
Wang et al. (2017) \cite{46} &  89.7\% &  70.6\% &  61.3\% &  78.2\%\\\cline{1-1}\cline{2-2}\cline{3-3}\cline{4-4}\cline{5-5}
Ma et al. (2018) \cite{47} &  - &  - &  76.9\%(JHMDB) &  -\\\cline{1-1}\cline{2-2}\cline{3-3}\cline{4-4}\cline{5-5}
Tu et al. (2018) \cite{48} &   94.5\% &   - &   69.8\% &  -\\\cline{1-1}\cline{2-2}\cline{3-3}\cline{4-4}\cline{5-5}
Yang et al. (2019) \cite{50} &   92.6\% &   - &   - &  -\\\cline{1-1}\cline{2-2}\cline{3-3}\cline{4-4}\cline{5-5}
Rashwan et al. (2020) \cite{61} &   78.43\% &   \underline{87.94\%} &  - &  -  \\\cline{1-1}\cline{2-2}\cline{3-3}\cline{4-4}\cline{5-5}
Avola et al. (2020) \cite{62} &   96.2\% & - &  72.1 &  -  \\\cline{1-1}\cline{2-2}\cline{3-3}\cline{4-4}\cline{5-5}
Khan et al. (2020) \cite{63} &  - &   - &  \underline{93.7\%}&  -  \\\cline{1-1}\cline{2-2}\cline{3-3}\cline{4-4}\cline{5-5}
Sharif et al. (2020) \cite{64} &  - &   - &  92.6\%&  \textbf{98.2\%} \\\cline{1-1}\cline{2-2}\cline{3-3}\cline{4-4}\cline{5-5}
Asghari et al. (2020) \cite{74} &  \underline{98.4\%} &   - &  84.2\%&  - \\\cline{1-1}\cline{2-2}\cline{3-3}\cline{4-4}\cline{5-5}
Majd et al. (2020) \cite{75} & 93.6\% &   - &  66.2\%&  - \\\cline{1-1}\cline{2-2}\cline{3-3}\cline{4-4}\cline{5-5}
\textbf{Proposed Method} &   \textbf{98.66\%} &   \textbf{91.32\%} &   \textbf{95.04\%} &   \underline{97.65\%} \\\cline{1-1}\cline{2-2}\cline{3-3}\cline{4-4}\cline{5-5}
\end{tabulary} 
\end{table*}\quad

\section{Conclusions}
In this study, we propose the 3DCNN-based multi human action recognition method. Compared with existing methods, our method can simultaneously detect and recognize actions made by many persons in the same video. For video surveillance systems, our approach enables human action recognition and summarization in public areas where people simultaneously act in the scene. With the proposed representation of data, which was performed prior to the recognition phase, our model accurately recognizes most actions. The advantage of the proposed approach is that recognition and summarization are for many people acting in the scenes. For YouTube and movie videos, in most methods, the videos are classified to extract the actions in them before summarizing these actions. Compared with the results of existing methods, our results are convincing.

\section*{Acknowledgment}

This publication was made by NPRP grant \# NPRP8-140-2-065 from the Qatar National Research Fund (a member of the Qatar Foundation). The statements made herein are solely the responsibility of the authors.

\end{document}